\documentclass{article}

\usepackage{microtype}
\usepackage{graphicx}
\usepackage{booktabs}
\usepackage{multirow}
\usepackage{multicol}
\usepackage{balance}
\usepackage{siunitx}
\usepackage{subcaption}
\usepackage{graphicx}
\usepackage[percent]{overpic}
\usepackage{xcolor}
\usepackage{caption}
\usepackage{hyperref}

\usepackage[accepted]{icml2025}

\usepackage{amsmath}
\usepackage{amssymb}
\usepackage{mathtools}
\usepackage{amsthm}

\usepackage[capitalize,noabbrev]{cleveref}

\theoremstyle{plain}

\theoremstyle{definition}

\theoremstyle{remark}

\usepackage[textsize=tiny]{todonotes}
\makeatletter
\def\icml@notice{}         
\def\@copyrightspace{}     
\makeatother

\icmltitlerunning{Mechanistic Interpretability of Antibody Language Models Using SAEs}

\begin{document}
\fontsize{9}{11}\selectfont

\twocolumn[
\icmltitle{Mechanistic Interpretability of Antibody Language Models Using SAEs}

{\centering
\small
Rebonto Haque$^{1}$\quad
Oliver M. Turnbull$^{1}$\quad
Anisha Parsan$^{2,3}$\quad
Nithin Parsan$^{2}$\quad
John J. Yang$^{2}$\quad
Anna L. Beukenhorst$^{4}$\quad
Charlotte M. Deane*$^{1}$\par\vspace{4pt}
\small
$^{1}$Department of Statistics, University of Oxford, UK\quad
$^{2}$Reticular, San Francisco, USA\quad
$^{3}$EECS, MIT, Cambridge MA, USA\quad
$^{4}$Leyden Laboratories BV, Leiden, The Netherlands\quad

\textit{*Correspondence:}
\href{mailto:deane@stats.ox.ac.uk}{deane@stats.ox.ac.uk}\par
}

\vskip 0.2in
]

\begin{abstract}
Sparse autoencoders (SAEs) are a mechanistic interpretability technique that have been used to provide insight into learned concepts within large protein language models. Here, we employ TopK and Ordered SAEs to investigate autoregressive antibody language models, and steer their generation. We show that TopK SAEs can reveal biologically meaningful latent features, but high feature–concept correlation does not guarantee causal control over generation. In contrast, Ordered SAEs impose a hierarchical structure that reliably identifies steerable features, but at the expense of more complex and less interpretable activation patterns. These findings advance the mechanistic interpretability of domain-specific protein language models and suggest that, while TopK SAEs suffice for mapping latent features to concepts, Ordered SAEs are preferable when precise generative steering is required.
\end{abstract}

\section{Introduction}
\label{submission}
Antibodies are a key part of the body’s adaptive immune response. They are characterised by their ability to bind to a specific antigen and subsequently neutralise it or initiate an immune response. Their extensive sequence—and therefore structural—diversity enables binding to virtually any target antigen \cite{chiu_antibody_2019}.  

The antigen-binding domain of antibodies is made up of variable heavy (VH) and variable light (VL) chains. Antibody-antigen binding specificity and affinity are largely determined by structural units known as complementarity-determining regions (CDRs), with each VH and VL chain having 3 CDRs, making a total of 6. Within our genome, there are numerous V, D, and J gene segments which together code for the VH and VL chains.

The combinatorial assembly of the discrete gene segments–V (variable), D (diversity), and J (joining) for the VH (heavy) chain, and V and J for the VL (light) chain–give rise to the final sequence diversity within the VH/VL chains.  Somatic hypermutations, characterised by random nucleotide substitutions occurring at rates considerably higher than the genomic background, in the joint V(D)J segment further increases sequence diversity \cite{andreano_immunodominant_2021}.

The ability to bind any target antigen with high specificity and affinity makes antibodies ideal candidates for drug discovery. As a result, antibody drugs hold a major and growing share of the total pharmaceutical market \cite{crescioli_antibodies_2025}. Antibody drug development pipelines need to identify candidates which bind specifically and with high affinity to the target antigen, while also being ‘developable’ \cite{jarasch_developability_2015}. ‘Developability’ refers to properties required for a successful drug such as immunogenicity, solubility, specificity, stability, manufacturability, and storability \cite{raybould_therapeutic_2022}. 

Machine learning models have been used to optimise multiple steps of antibody-drug development pipelines from library generation \cite{turnbull_p-iggen_2024-1, shuai_iglm_2023, nijkamp_progen2_2023} and light chain generation \cite{capel_lichen_2026} to humanisation during lead optimisation \cite{chinery_humatch_2024, prihoda_biophi_2022, marks_humanization_2021}. In this work we will primarily use p-IgGen, a GPT-like decoder-only antibody language model trained on antibody-sequence data, consisting of ~17M parameters \cite{turnbull_p-iggen_2024-1}. The authors released a paired model, as well as a finetuned version capable of generating diverse antibody libraries with developable properties. 

The lack of interpretability of machine learning models contributes to a lack of trust in model predictions, difficulty determining whether biologically relevant features are being used to make predictions and difficulty detecting overfitting. Collectively, these pose a barrier when employing language models for drug discovery \cite{vamathevan_applications_2019}. SAEs offer a promising approach to identify human-interpretable concepts learned by models and steer their generation \cite{templeton_scaling_2024}. Prior works have used SAEs to understand the inner mechanisms of protein language models (PLMs) \cite{adams_mechanistic_2025,parsan_towards_2025-1,simon_interplm_2024}, and steer model output. However, to date, SAEs have not been used to interrogate autoregressive protein or antibody-specific language models. 

In this work we aim to advance the interpretability of antibody language models, using SAEs to identify biologically relevant features of interest learned by p-IgGen, and predictably steer its generation. We identify antibody-specific features, such as the complementarity-determining region (CDR) identity and germline gene identity, and use them to steer p-IgGen generation for specific germline gene identities. While p-IgGen is our primary mechanistic case study, we additionally evaluate whether the interpretability–steerability patterns we observe are specific to a single architecture. To this end, we replicate key analyses on two additional antibody language models—IgLM \cite{shuai_iglm_2023} and ProGen2-OAS \cite{nijkamp_progen2_2023}—using the same SAE-based pipeline. These experiments are not intended as exhaustive model comparisons, but as evidence that the qualitative conclusions we draw from p-IgGen generalize across scale and architecture.

Overall, we demonstrate the applicability of SAEs for incorporating rational design principles to antibody library generation. We show that TopK SAEs can accurately identify interpretable latents underpinning model generation, whereas Ordered SAEs can identify steerable features capable of tuning model generation.

\section{Related Work}
\subsection{Mechanistic Interpretability}
Mechanistic interpretability refers to the approach of explaining complex machine learning systems through the behaviour of their functional units \cite{kastner_explaining_2024} by decomposing or reverse-engineering systems into their more elementary computations \cite{rai_practical_2025}. The eventual goal is to discover causal relationships between model inputs and corresponding outputs. 

Within the context of transformer-based language models, there are three main ideas relevant for mechanistic interpretability research: features, circuits and universality \cite{rai_practical_2025}. Features refer to human-interpretable properties that are encoded by model activations \cite{templeton_scaling_2024}. Circuits inform how these features are extracted from model inputs and processed to influence model outputs \cite{olah_zoom_2020}. Finally, universality determines whether features and circuits identified for a specific model and task exist in other models and tasks \cite{olah_zoom_2020}. This paper specifically focuses on the identification of features from language models and using these features to steer model generation.

\subsection{Sparse Autoencoders}
Sparse Autoencoders (SAEs) have specifically been employed in mechanistic interpretability for feature discovery. They are able to tackle the issue of feature superposition resulting in polysemantic neurons, where any given neuron encodes multiple, often unrelated features. SAEs overcome this problem by projecting dense neuron activations into a sparser latent space using a sparse encoder, Equation \ref{eq:encoder}, whilst ensuring the latent representation can be reconstructed back into the original neuron representation by a decoder following sparsification, Equation \ref{eq:decoder}. 
\begin{equation}
  z = g(ReLU\bigl(W_{\text{enc}}\,x + b_{\text{enc}}\bigr))
  \label{eq:encoder}
\end{equation}
\begin{equation}
  \hat{x} = W_{\text{dec}}\,z + b_{\text{dec}}
  \label{eq:decoder}
\end{equation}

where \(W\) are the weight matrices and \(b\) are the bias vectors, \(enc\) and \(dec\) denote the encoder and decoder respectively,

\(x\) is the original hidden representation,  
\(z\) the latent representation,  
and \(\hat{x}\) the reconstructed hidden representation. \(ReLU\) activation is applied to the latent representation following encoding and \(g\) is a sparsification function. 

\subsubsection{TopK SAEs} TopK SAEs \cite{gao_scaling_2024} limit the number of active latents to \(k\), where \(k \ll d_{\mathrm{in}} \ll d_{\mathrm{sae}}\).  \(d_{\mathrm{in}}\) is the input hidden dimensions,  and \(d_{\mathrm{sae}}\) is the latent or dictionary dimensions. Equation \ref{eq:loss} shows the loss computation.
\begin{equation}
  L(x) 
  = \underbrace{\|x - \hat{x}\|_2^2}_{\text{Reconstruction loss}}
    \;+\;
    \underbrace{c}_{\text{Sparsity constraint}}
  \label{eq:loss}
\end{equation}
The \(L(x)\) reconstruction loss compares the decoded representation \(\hat{x}\) with the original hidden representation \(x\).  When a sparsification function is not directly applied during encoding, a separate sparsity constraint is added in loss computations, which is usually a variation of an L1 regularisation loss \cite{zhang_sparse_2018}. 

\subsubsection{Ordered SAEs (O-SAEs)} Ordered SAEs \cite{wang_enforcing_2025} follow a nested SAE architecture, enabling hierarchical ordering of SAE latents. Importantly, compared to the traditional TopK SAE architecture which arbitrarily orders hierarchical latents within the dictionary space, O-SAEs enforce a strict, consistent, hierarchical ordering of latents. This is because TopK SAEs enforce sparsity within the entire dictionary space in one go, whereas O-SAEs follow a nested approach and effectively train a number of individual, nested SAEs which occupy an increasing portion of the dictionary space.

O-SAEs introduce two core components: (i) \emph{per-index nested grouping}, and (ii) \emph{strictly decreasing truncation weights} in order to ensure consistent ordering. 

(i) For each truncation level $m \in \{1,\dots,d_{\mathrm{sae}}\}$, the first $m$ rows of the encoder and decoder are isolated:\begin{equation}
  W_{\mathrm{enc}}^{(m)} = [W_{\mathrm{enc}}]_{1:m,\,:}, 
  \quad
  W_{\mathrm{dec}}^{(m)} = [W_{\mathrm{dec}}]_{1:m,\,:}
\end{equation}
In Eq.~(4) the encoder–decoder pair \(\bigl(W_{\text{enc}}^{(m)},\, W_{\text{dec}}^{(m)}\bigr)\) re-uses the first \(m\) rows of the full weight matrices. Because every smaller autoencoder is a strict subset of the larger one, any latent \(i \le m\) is shared across all groups that follow. This “per-index nested grouping” forces early latents to model global structure that remains useful for every deeper stage. Per-index grouping ensures non-random sampling of dictionary sizes, unlike in Matryoshka SAEs \cite{bussmann_learning_2025}, increasing the overall consistency of results. 

(ii) Each partial reconstruction is weighted by a \emph{monotonically decreasing} probability $p_M(m)$, so that early (low-index) features incur a higher penalty when failing to capture coarse structure. The per-truncation loss is
\begin{equation}
  L_m(x) = p_M(m)\,\left\lVert x - W_{\mathrm{dec}}^{(m)\top}\,W_{\mathrm{enc}}^{(m)}\,x \right\rVert_2^2
\end{equation}
and summing over all $m$ promotes the model to learn the most “abstract” elements first, with progressively finer details later. Combining the decreasing probability weights with nested latents further enforces ordering of identified latents and maintains a stricter hierarchy. 

\section{Methods}
\subsection{Sparse Autoencoder Training}
We adapted TopK Sparse Autoencoders (SAEs) from the EleutherAI/sparsify GitHub repository
(\url{https://github.com/EleutherAI/sparsify}). Ordered SAEs were implemented based on the paper by \cite{wang_enforcing_2025} and related repositories \cite{marks2024dictionary_learning}. Data and code are available at: \url{https://tinyurl.com/4ebamp4f}

\subsubsection{Base models and activation extraction}
Our primary experiments use p-IgGen, an autoregressive antibody language model trained on paired OAS. To test robustness, we also run the same SAE training and concept-probing pipeline on IgLM and ProGen2-OAS, which differ in architecture and scale. We trained both the TopK and Ordered SAEs on hidden layer activations of respective models, generated from the original p-IgGen training set \cite{turnbull_p-iggen_2024-1}. For p-IgGen, we extract activations from each of its 4 hidden layers. For IgLM and ProGen2-OAS we only extract activations from their final layers.

The training set contained \textbf{1,800,545} VH/VL paired sequences from the Observed Antibody Space database (OAS) \cite{olsen_observed_2022, kovaltsuk_observed_2018}. When generating p-IgGen activations for training, we concatenated the paired VH and VL sequences together, with appropriate start and end tokens added, and passed them into p-IgGen to generate hidden activations. This generated 4 sets of hidden activations, one from each hidden layer. For Ordered SAE training, we randomly subsampled \textbf{100,000} sequences to decrease training time.

We further trained SAEs on final hidden layer activations of IgLM and ProGen2-OAS using the same p-IgGen training sets as before. Since IgLM and ProGen2-OAS were originally trained using unpaired sequences, we generated activations for unpaired VH and VL chains with appropriate start, end, and padding tokens from the respective vocabularies. 

\subsubsection{TopK SAE}
The model's input dimensions $d_{\mathrm{in}}$ were projected onto a higher-dimensional latent/dictionary size $d_{\mathrm{sae}}$, where  $d_{\mathrm{sae}} = d_{\mathrm{in}}\times r = d_{\mathrm{in}}\times 32$. $r = 32$ is the expansion factor. ReLU activation was applied to the projection, $z = \operatorname{ReLU}\bigl(W_{\mathrm{enc}}\,x + b_{\mathrm{enc}}\bigr)$, followed by a Top-$k$ sparsification with $k = 32$, retaining only the top 32 activations by magnitude. The resulting dictionary size was \textbf{32x}. Decoder weights were initialised as the unit-normalised transpose of the encoder weights to stabilise training. Training used a batch size of 8 and Adam optimisers throughout, with a custom learning rate $\eta = \frac{2 \times 10^{-4}}{\sqrt{\,d_{\mathrm{sae}} / 16{,}384\,}}$.

\subsubsection{Ordered SAE}
Ordered Sparse Autoencoders (O-SAEs) were adopted to retain higher-level, abstract features within our latent space and hierarchically arrange the latents.
In our setup, we used expansion factor $r=8$, yielding a dictionary size $d_{\mathrm{sae}} = d_{\mathrm{in}}\times r =d_{\mathrm{in}}\times 8$. All models were trained with Adam optimisers at a fixed learning rate $\eta = 1\times10^{-4}$.  We chose a smaller maximum dictionary size for the O-SAEs to speed up training, effectively reducing the total number of nested SAEs being trained. Due to per-index grouping, O-SAEs need to train several nested SAEs based on the total dictionary size, whereas the regular TopK architecture only trains a single model. 

For p-IgGen, we initially used Batch TopK activation for sparsification, with $k=32$ (following an earlier version of \cite{wang_enforcing_2025}), where activations across all residues in a batch are pooled and the top $k \times B$ selected globally ($B$ = batch size). This maintains $k$ active latents on average, but allows per-residue sparsity to vary. Subsequently, IgLM and ProGen2-OAS used standard TopK (fixed $k$ per residue) for fairer comparison with TopK SAEs.

\subsection{Targeted Feature Identification using SAEs}
\subsubsection{Training data}
Paired antibody sequence data were obtained from OAS, Coronavirus Antibody Database (CoV-AbDab) \cite{raybould_cov-abdab_2021}, and the Patent and Literature Antibody Database (PLAbDab) \cite{abanades_patent_2024}. A total of 149,069 sequences were obtained from the respective datasets, based on their binding specificities to SARS-CoV2 RBD (binder and non-binder).  

The data was clustered based on CDR sequence similarity using CD-HIT \cite{li_clustering_2001}, with a 0.8 similarity threshold on the total CDR sequence.  The clusters were then randomly split into the training-validation-test set, whilst ensuring members of the same cluster were in only one of the three possible splits. The splits were further stratified based on binding specificity to SARS-CoV2 RBD. 

This specific dataset was originally prepared for a separate project, and the SARS CoV2 RBD binding properties of the antibodies are not relevant for this study. Qualitatively, a dataset of equivalent size randomly sampled from OAS should produce the same results. 

The following concepts were studied to identify associated latents: CDR identity, which refers to whether a given residue lies within a specific CDR region, and V/J gene identity, which refers to the germline V or J gene segment was used to code for the final antibody sequence. For the CDR-identity, the training matrix was the latent activations for each residue. The CDR identity dataset had 7 classes (6 CDR identities and non-CDR regions). For sequence-level concepts, V/J gene identity, the training matrix was the mean pool of the \emph{non-zero} residue-level latent activations in a given sequence. 

\subsubsection{Linear Probe}
We trained a logistic regressor to act as a linear probe on the training-validation data. A logistic regressor (LR) was trained, employing 3-fold cross-validation grid search to optimise hyperparameter C. In logistic regression, C is the inverse of the regularisation strength: larger C applies less regularisation and can overfit, while smaller C applies more regularisation and can improve generalisation. Cross-validation was done during training by randomly shuffling and splitting the training data into 3 cross-validation sets. We selected the top 500 latents based on their positive logistic regression coefficients.

\subsubsection{Latent Selection}\label{latent_selection}
Latents were further validated on the validation set. Based on the strategy by Simon and Zou \cite{simon_interplm_2024}, the latent activations across the validation set were normalised using MinMax scaling. In this instance, similar to Simon and Zou, we took the mean activation of the latents across all residues in the sequence. For each normalised latent, binary latent-on/latent-off labels using activation thresholds of 0.1, 0.2, 0.5, 0.8, 0.9 were applied. For each latent--concept pair, we defined a latent as an interpretable feature only if it achieved an $F_1$ score that exceeded a class-prevalence baseline by a fixed margin. Let $\pi$ denote the prevalence of the positive class for the concept in the validation set. We first computed the $F_1$ score of a trivial always-positive classifier,
\begin{equation}
F_{1,\mathrm{all+}} \;=\; \frac{2\pi}{1+\pi}.
\end{equation}
A latent was labelled a feature if, for any tested activation threshold, its $F_1$ satisfied
\begin{equation}
F_1 \;>\; F_{1,\mathrm{all+}} + 0.2
\end{equation}
This criterion accounts for class imbalance and requires each latent to provide predictive signal beyond a strong non-informative baseline, rather than exploiting the marginal class distribution. It is worth noting that the threshold for "feature" is somewhat arbitrary. Latents just below the threshold are likely to still have meaningful correlations to a target concept.

\subsubsection{Antibody sequence alignment}
Antibody sequences were aligned using ANARCI \cite{dunbar_anarci_2016} and the IMGT numbering \cite{lefranc_imgt_2003}.

\subsection{Steering}
Steering was implemented based on the strategy by Templeton et al.\ \cite{templeton_scaling_2024}. Each latent can be represented by its corresponding decoder vector $d^{(i)} = W_{\mathrm{dec}}[i, :]$, where $d^{(i)}$ is the decoder vector for latent $i$ and $W_{\mathrm{dec}}$ is the decoder weight matrix. Steering is performed by scaling the decoder vector and adding it to the original hidden state (Equation \ref{eq:steering}).
\begin{equation}
h_l^{*} \;\leftarrow\; h_l \;+\; \alpha \cdot d^{(i)}
\label{eq:steering}
\end{equation}
Here, $\alpha$ is the steering factor and $h_l$ is the hidden state before the intervention and $h_l^{*}$ is the hidden state following the intervention. 

\section{Results}

\subsection{TopK SAE-identified Features are Interpretable, Antibody-specific Concepts, but Not Steerable}

\subsubsection{TopK SAE latents preserve biological information following sparsification}
TopK sparsification represents each token with far fewer latents compared to the hidden neurons, creating a possibility for information loss during sparsification. To qualitatively check whether antibody-specific information is retained within the latents, we compared the accuracies of using latents and hidden neurons on residue and sequence-level property prediction tasks. 

We began the analysis by focusing on TopK SAEs trained on the final layer activations (layer 3), as seen in previous works \cite{parsan_towards_2025-1}, since they contain the most complete representation of the sequence. Logistic regressor training on CDR identities using latent activations obtained a validation accuracy of \textbf{0.99}, while using hidden neuron activations resulted in a validation accuracy of \textbf{0.98}. This showed that residue-level CDR identity information is preserved in the activated latents. 

In order to check sequence-level features, we investigated germline gene predictions. We focused on heavy J genes for simplicity, given they have less allelic variation compared to heavy V genes. LR training resulted in a validation $F_1$  macro score of \textbf{0.93}. We used an $F_1$  score due to significant class imbalance in heavy J genes within the training and validation sets. Table \ref{tab:performance} reports the precision, recall and $F_1$ -scores of the different gene identities studied and shows how the model accurately predicts each IGHJ class using the SAE latents. Overall, the high $F_1$  macro score shows how the activated latents also preserve sequence-level features. 

Our results indicate that SAE latents collectively represent antibody information following sparsification when applied to an antibody language model, similar to results seen in previous works for general protein language models \cite{simon_interplm_2024, adams_mechanistic_2025, parsan_towards_2025-1}. We therefore carried out further analysis of SAE latents to investigate their overall interpretability.
\begin{table}[ht]
  \centering
  \caption{Precision, recall, and $F_1$-score per IGHJ class.}
  \label{tab:performance}
  \begin{tabular}{lccc}
    \toprule
    Class & Precision & Recall & $F_1$-score \\
    \midrule
    IGHJ1 & 0.91 & 0.71 & 0.80 \\
    IGHJ2 & 0.94 & 0.93 & 0.94 \\
    IGHJ3 & 0.96 & 0.96 & 0.96 \\
    IGHJ4 & 0.98 & 0.99 & 0.98 \\
    IGHJ5 & 0.94 & 0.95 & 0.95 \\
    IGHJ6 & 0.98 & 0.97 & 0.98 \\
    \midrule
    Macro average & 0.95 & 0.92 & 0.93 \\
    \bottomrule
  \end{tabular}
\end{table}

\subsubsection{TopK latent activations are visually interpretable}
To investigate whether SAE latents provide a tangible method of interpreting the logic behind model generation, we compared the activation patterns of latents and neurons correlated to properties of interest. As a baseline, we compared activations of the top correlated latents and neurons for CDRH3. 

\begin{figure*}[t]
  \centering
  \begin{subfigure}[t]{0.98\textwidth}
    \centering
    \includegraphics[width=\linewidth]{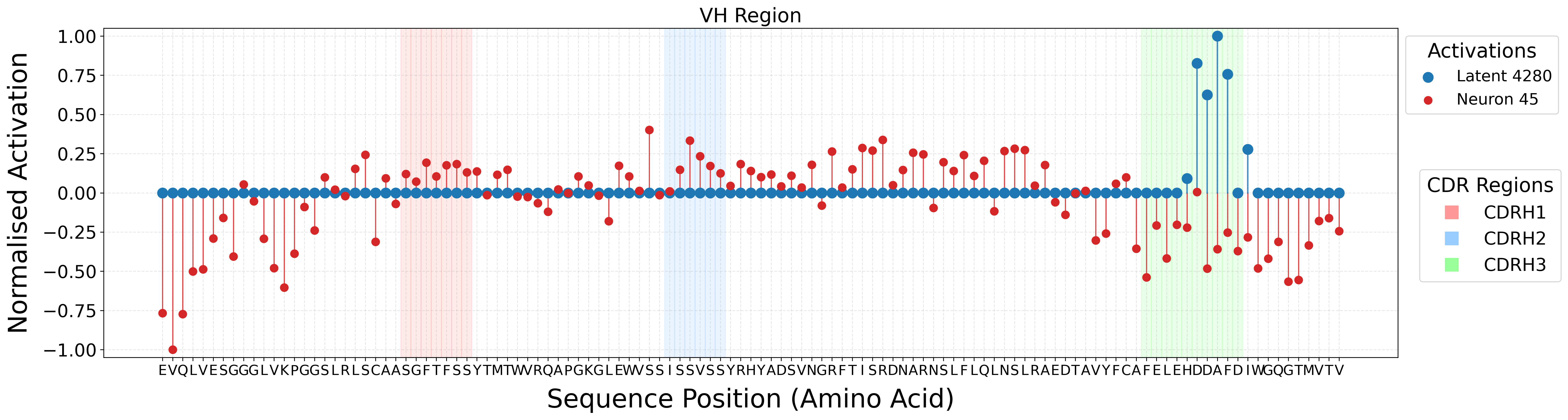}
    \subcaption{Latent (blue) vs neuron (red) activations for CDRH3 identity}
    \label{fig:neuron-vs-latent_a}
  \end{subfigure}

  \begin{subfigure}[t]{0.98\textwidth}
    \centering
    \includegraphics[width=\linewidth]{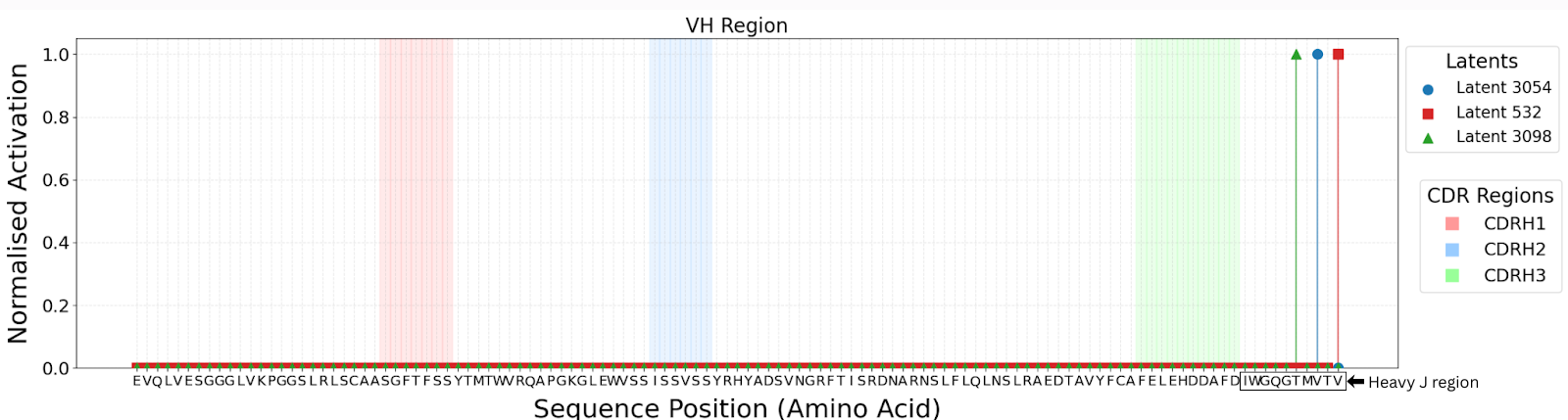}
    \subcaption{Latent activations for IGHJ3}
    \label{fig:neuron-vs-latent_b}
  \end{subfigure}

  \caption{TopK latent activations (blue) and hidden neuron activations (red) for CDRH3 (a), and TopK latent activations for IGHJ3 (b). The x-axis shows the amino-acid sequence of the VH region of a test antibody; the y-axis shows normalised activation. CDRs are coloured CDRH1 (red), CDRH2 (blue), and CDRH3 (green). Latent activations localise to the expected regions—CDRH3 in panel (a) and the heavy J region in panel (b), whereas neuron activations are scattered across the sequence with no discernible pattern.}
  \label{fig:neuron-vs-latent}
\end{figure*}

Visual investigation showed that latent activations are sparse and specific to CDRH3 residues, compared to neurons which activate across the sequence without any immediately recognisable pattern (Figure \ref{fig:neuron-vs-latent}). This may be explained by the polysemanticity of neurons, where multiple features specific to several unrelated residues are represented by the same neuron. When investigating activation patterns, this complicates using neurons as a tool for interpretability and highlights the potential greater explainability of SAE-derived latents. 
\captionsetup{font=small,skip=6pt} 

In order to investigate the utility of SAE activations for the mechanistic interpretability of sequence-level concepts, as opposed to residue-level concepts, we investigated heavy J gene activations as sequence-level concepts. Similar to residue-level observations, latents corresponding to heavy J gene identity activate on residues representing the concept, i.e. gene identity. In this instance, the top correlated latents were activated on the J domain of examined antibody sequences. Sequence-level representations are mean pools of the original residue-level representations, leading to an overall loss of positional information. Therefore, the top correlated latents also encode intrinsic positional information (Figure \ref{fig:neuron-vs-latent}b). This provides an opportunity to identify the residues responsible for a global, sequence-level feature, with potential implications for the understanding of the sequence and structural basis of antibody properties.

To quantify the predictive properties of our identified features, we carried out an activation-threshold analysis (Supplementary Table \ref{tab:ighj-features}, Methods 3.2.3).

%

\subsection{Ordered SAEs Identify More Steerable Features Compared to TopK SAEs}
\subsubsection{p-IgGen generation can be predictably steered using Ordered SAE latents}
\begin{figure*}[t]
    \centering
    \begin{subfigure}[t]{0.48\linewidth}
        \centering
        \includegraphics[width=\linewidth]{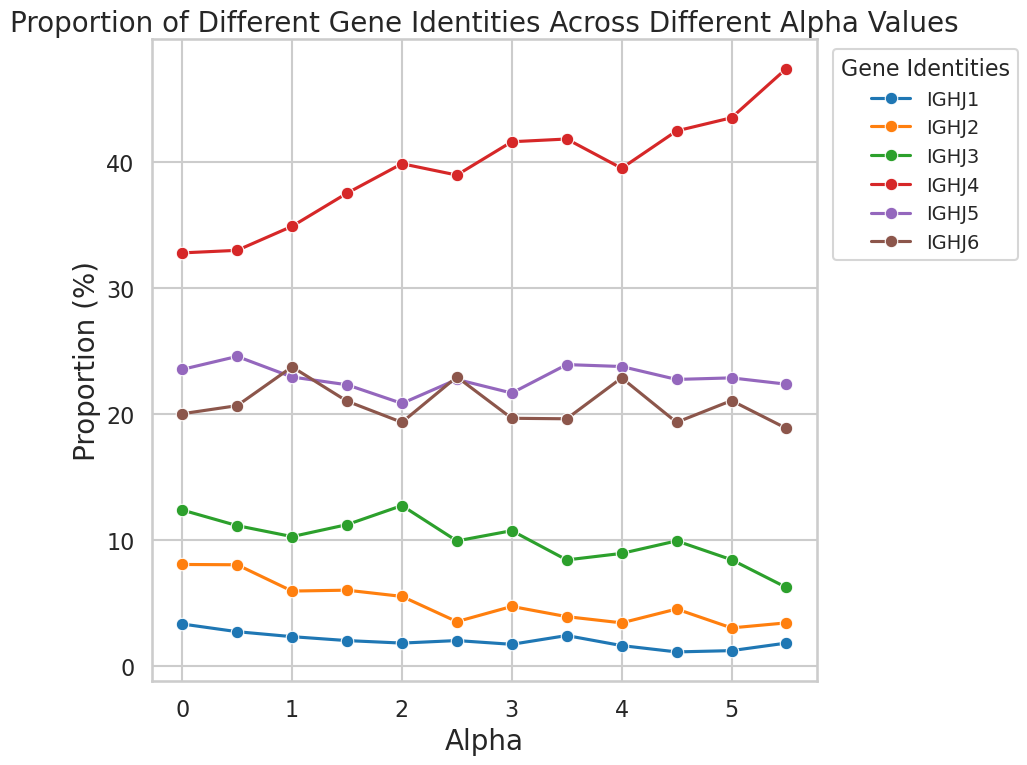}
        \subcaption{Positive steering}
        \label{fig:Successful_steering_a}
    \end{subfigure}\hfill
    \begin{subfigure}[t]{0.48\linewidth}
        \centering
        \includegraphics[width=\linewidth]{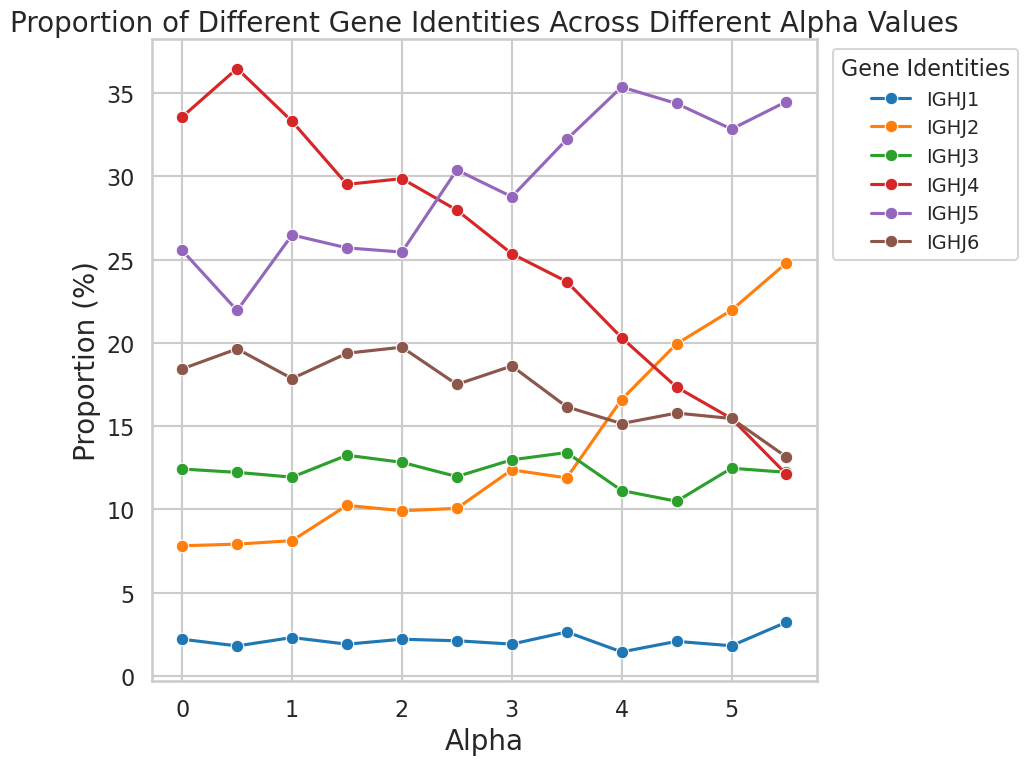}
        \subcaption{Negative steering}
        \label{fig:Successful_steering_b}
    \end{subfigure}

    \vspace{-2mm}
    \caption{Results of IGHJ4 steering using O-SAE p-IgGen latent 12 (a) and 49 (b). Y-axis shows the proportion of generated sequences. Plots are coloured by heavy J gene identity. X-axis shows the steering factor used (alpha). Results are for a library of 1000 p-IgGen-generated sequences. Latent 12—positively associated with IGHJ4 based on activation-threshold analysis—increases IGHJ4 proportion under positive steering, whereas latent 49—which has a strongly negative logistic regression coefficient—decreases IGHJ4 under the same steering.}
    \label{fig:Successful steering}
\end{figure*}

In addition to assessing how well latents predict the target concept, we also tested feature steering as an indicator of feature importance \citep{parsan_towards_2025-1}: if increasing a latent consistently steers generation in a desired direction, the corresponding feature is likely important for that direction. We were unable to successfully steer on TopK latents within p-IgGen's latent space (See Supplementary \ref{case_study}), which may be attributed to known issues within these architectures such as feature splitting and absorption \citep{chanin_is_2024}. We subsequently evaluated Ordered SAEs, which build a hierarchical latent space that preserves both high-level and fine-grained features \citep{bussmann_learning_2025}, but result in a less interpretable localisation pattern (Supplementary Figure \ref{fig:nested latent activation}).

We conducted a linear probe and subsequent activation-threshold analysis (Methods \ref{latent_selection}) to identify features correlated to IGHJ4 in layer 3. Due to the implicit hierarchy in features, we ranked latents based on their dictionary index, with smaller indices representing higher-level features. Additionally, to prioritize these higher level features, we carried out activation-threshold analysis on the first 50 latents based on dictionary index. We identified latent 12, which was positively associated with IGHJ4 based on activation-threshold analysis, and latent 49, which had a strongly negative logistic regression coefficient (Figure \ref{fig:Successful steering}).

Positively steering on latent 12 increased IGHJ4 proportion in model generation (Pearson's R = 0.943, p = \num{4.373e-06} and Spearman's correlation = 0.944, p = \num{3.927e-06}). Conversely, positively steering on latent 49 decreased IGHJ4 proportion in model generation (Pearson's R of -0.980, p-value = \num{2.525e-08} and Spearman's correlation = -0.986, p = \num{4.117e-09}).

To investigate the features, we plotted their activations on specific IMGT positions across a random set of IGHJ4 containing sequences (Supplementary Figure \ref{fig:nested latent activation}) The latents activate more broadly across the sequence than TopK latents. One reason for this may be the abstraction which enables steerability. The latents inform downstream generation, and therefore signal early on during sequence generation about the identity of the J-region. Given the latents communicate longer-range concepts, it is possible that latent activation is not linked to any specific residue, rather a range of residues within a defined length of the sequence.

\subsubsection{Nested steering generates novel, diverse, antibody-like sequences}
To assess whether nested steering alters the generative distribution while retaining antibody-like properties, we evaluated steered libraries using sequence novelty (Hamming distance to the held-out test set), diversity (nearest-neighbour cosine distance), and protein-likeness (ESM-2 likelihood \cite{lin_evolutionary-scale_2023}).
Across all metrics, steered sequences remained close to the natural baseline, with only modest absolute shifts (Table~\ref{tab:steered_generation_analysis}).

Steered sequences using O-SAE latent 12 (Figure \ref{fig:Successful steering} (a)) and alpha 6.50 were slightly more distant from the test set than natural sequences for both VH (23.83 vs.\ 22.91; $\Delta\approx 0.92$) and VL (20.34 vs.\ 19.81; $\Delta\approx 0.53$), indicating increased novelty.
Steering also produced a small increase in diversity (0.21 vs.\ 0.20; $\Delta\approx 0.01$)
Protein-likeness was essentially unchanged, with ESM-2 likelihoods differing by only $\sim$0.005 ($-0.325$ vs.\ $-0.330$), suggesting steered sequences remain on a highly plausible antibody manifold. ANARCI \cite{dunbar_anarci_2016} could number all 1000 of the generated sequences and distinguish between the heavy and light chains, further indicating antibody-likeness.

For statistical testing, we compared the distribution of each metric between steered and natural sequences using a two-sided Mann--Whitney $U$ test.
Although all comparisons were statistically significant ($\textit{P}<0.05$), effect sizes were uniformly small: rank-biserial correlations were $<0.2$ for each metric, indicating substantial overlap between the steered and natural distributions.

\begin{table}
\centering
\caption{Steered generation analysis. Asterisks indicate a statistically significant difference between steered and natural sequences under a two-sided Mann--Whitney $U$ test ($p<0.05$).}
\label{tab:steered_generation_analysis}
\begin{small}
\setlength{\tabcolsep}{4pt}
\begin{tabular}{p{0.46\columnwidth}ccc}
\toprule
\textbf{Metric} & \textbf{Steered} & \textbf{Natural} & \textbf{$\Delta$} \\
\midrule
Mean VH Hamming dist.\ to test\ set & 23.83$^{*}$ & 22.91 & 0.92 \\
Mean VL Hamming dist.\ to test\ set & 20.34$^{*}$ & 19.81 & 0.53 \\
Diversity         & 0.21$^{*}$  & 0.20  & 0.01 \\
ESM2-likelihoods                    & $-0.325^{*}$ & $-0.330^{*}$ & 0.005 \\
\bottomrule
\end{tabular}
\end{small}
\end{table}

\subsection{SAE-Based Interpretability Transfers Across Antibody Language Models}
In order to demonstrate the generalisability of SAEs for mechanistic interpretability studies, we examined two additional antibody language models of different scale and architecture:
(i) \textbf{IgLM} (4 layers, 512 hidden dimensions, 13M parameters) and
(ii) \textbf{ProGen2-OAS} (27 layers, 1536 hidden dimensions, 764M parameters).

Using the TopK pipeline on final-layer activations, we identified SAE latents that are maximally predictive of IGHJ germline identity. High-scoring features were found for IGHJ1 (ProGen2-OAS: max $F_1 = 0.535$), IGHJ3 (IgLM: max $F_1 = 0.641$; ProGen2-OAS: max $F_1 = 0.775$), IGHJ4 (IgLM: max $F_1 = 0.692$; ProGen2-OAS: max $F_1 = 0.679$), and IGHJ6 (IgLM: max $F_1 = 0.773$; ProGen2-OAS: max $F_1 = 0.692$). In contrast, no strong TopK features were detected for IGHJ2 or IGHJ5.

Similar to p-IgGen latents, IgLM and ProGen2-OAS TopK latents correlated to IGHJ4 activate at or near the J-region of antibodies (Supplementary Figure \ref{fig:IgLM_ad_progen2_activations}). None of the top IgLM or ProGen2-OAS IGHJ features demonstrate a strong correlation to aligned positions as their corresponding p-IgGen features. This may be explained by low feature consistency with TopK SAEs \cite{wang_enforcing_2025}, where the nature of features identified is sensitive to the architecture of the base language model.

\begin{figure}[t]
    \centering

    \begin{subfigure}[t]{\columnwidth}
        \centering
        \includegraphics[width=\linewidth]{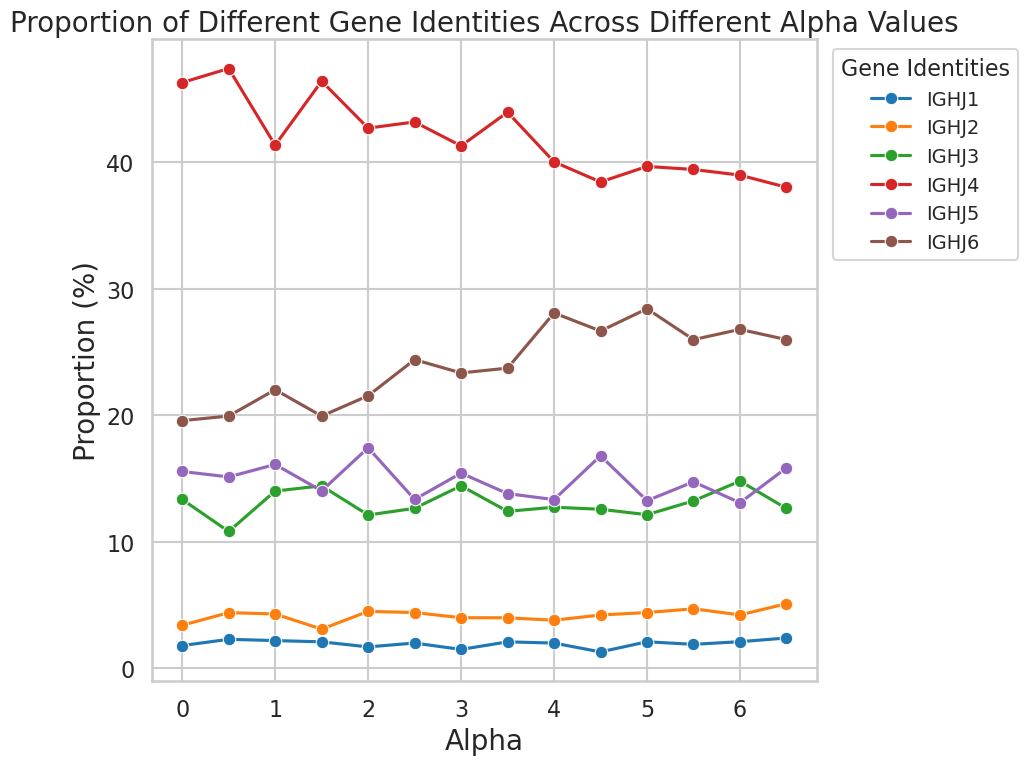}
        \subcaption{IgLM steering}
    \end{subfigure}\vspace{-2pt}

    \begin{subfigure}[t]{\columnwidth}
        \centering
        \includegraphics[width=\linewidth]{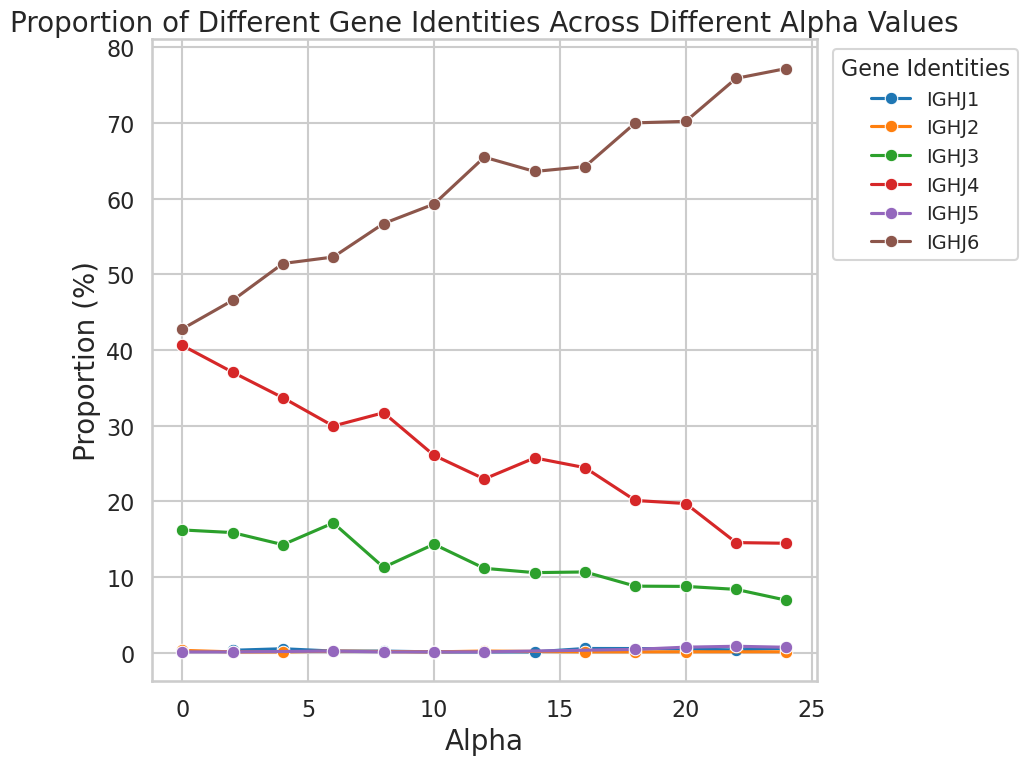}
        \subcaption{ProGen2-OAS steering}
    \end{subfigure}

    \vspace{-6pt}
    \caption{Results of IGHJ feature steering for O-SAE IgLM latent, positively correlated to IGHJ6 (a), and ProGen2-OAS latent, positively correlated to IGHJ6 (b). The y-axis shows the proportion of generated sequences. Plots are coloured by heavy J gene identity. The x-axis shows the steering factor used ($\alpha$). Results are for a library of 1000 generated sequences.}
    \label{fig:iglm_progen2_steering}
    \vspace{-3pt}
\end{figure}

Finally, to assess whether steering interventions transfer across model architectures, we identified Ordered SAE features strongly correlated with specific IGHJ identities. For both IgLM and ProGen2-OAS, we positively steered on a low-index feature associated with IGHJ6 (Figure~\ref{fig:iglm_progen2_steering}).

For ProGen2-OAS steering, correlations between the target IGHJ6 score and the steered latent activation were highly significant (Pearson $r = 0.986$, $p = 5.49\times 10^{-10}$; Spearman $\rho = 0.984$, $p = 1.611\times 10^{-9}$). For IgLM steering, we likewise observed significant correlations, though with smaller magnitude (Pearson $r = 0.872$, $p = 4.811\times 10^{-5}$; Spearman $\rho = 0.852$, $p = 1.077\times 10^{-4}$).

Replicating key experiments on IgLM and ProGen2-OAS suggests that two qualitative findings are robust: (i) SAE latents can recover antibody-relevant gene-identity structure from internal activations, and (ii) Ordered SAEs reliably expose steerable features. These results support the view that SAE-based interpretability is not confined to a single architecture or scale, but can serve as a general-purpose diagnostic tool across diverse antibody language models. 

\section{Conclusions and Future Outlook}
We show that SAEs can successfully be used to evaluate autoregressive antibody language models and identify learned domain-specific features. Current SAE implementations on PLMs for identifying features are limited by the assumption that feature correlation is equal to biological concept causation. Analysis of activated latents reveals high predictive performance does not always correspond to steerability. 

We found that some latents correlated to gene identity simply mark conserved positions without a clear causal link to gene identity. Some highly correlated features may highlight non-specific aspects of a concept, but be correlated to more concrete features. When we attempted to steer p-IgGen by amplifying these individual latents, the resulting antibody libraries did not show a reliable increase in IGHJ4 usage, further underscoring that high predictive power alone does not guarantee steerability. 

Ordered SAEs solve this by identifying hierarchical features which correspond to more abstract and higher-level concepts rather than simple residue-level identity. However, this comes at the cost of intuitive interpretability of activation patterns.  

One of the major limitations of employing SAEs to antibody language models is the lack of labelled datasets, unlike general proteins \cite{suzek_uniref_2007}. This prevents automated feature identification and annotation, which has allowed for their easy application for general protein language models (PLMs) \cite{simon_interplm_2024}. Lack of functionally annotated data remains a problem within the domain of antibody language models.

Several issues remain to enable the use of SAEs for PLM interpretability. For example, exhaustive steering analyses should be conducted to quantitatively measure the steerability of TopK and Ordered SAE features and SAE training should be scaled using more expansive, annotated datasets like FLAb \cite{chungyoun_flab_2024} to enable models to identify domain-specific abstractions and facilitate targeted manipulation of generative outputs. This would enable the generation of antibody libraries with properties such as developability and specificity, improving modern library generation methods and optimising the antibody drug development pipeline. 

\clearpage            
\balance
\section{Data Availability}
All the relevant training, validation and testing datasets, as well as relevant code used for analysis can be accessed through the private Zenodo link: \url{https://tinyurl.com/4ebamp4f}

\bibliographystyle{icml2025}
\balance
\bibliography{Part-2}



\newpage 
\appendix
\section{Supplementary Material}
\addcontentsline{toc}{section}{Appendix}
\subsection{Case Study of Heavy J Gene Identity for IGHJ4} \label{case_study}
The functional importance of SAEs for studying LLMs lies in their ability to interrogate specific, domain-relevant concepts rather than an undefined set of all possible ones. Here, we examine gene identity, as it directly influences antibody binding affinity and specificity \cite{deng_allelic_2025}. Specifically, we pick IGHJ4 genes for our analysis, due to their widespread clinical significance underpinned by the fact that they are the most widely utilised J genes in our immune repertoire. That is, the majority of heavy chain antibodies within any given individual originate from the IGHJ4 germline gene. 
\begin{table}[ht]
  \centering
  \caption{IGHJ4 TopK feature statistics across p-IgGen layers. A latent is counted as a feature if it satisfies $F_1 > F_{1,\mathrm{all+}} + 0.2$ for any tested activation threshold.}
  \label{tab:IGHJ4_layer_performance}

  \vspace{0.5em} 

  \begin{tabular}{lccc}
    \toprule
    IGHJ & Layer & Features & Max $F_1$‐score \\ \\
    \midrule
    \multirow{4}{*}{IGHJ4} 
           & Layer 0 & 1                   & 0.930       \\
           & Layer 1 & 4                   & 0.949       \\
           & Layer 2 & 4                   & 0.930       \\
           & Layer 3 & 1                   & 0.752       \\
    \bottomrule
  \end{tabular}
\end{table}

For our analysis, we decided to focus on the final layer (Table \ref{tab:IGHJ4_layer_performance}). 

\begin{figure*}[t]
  \centering

  \begin{subfigure}[t]{0.48\textwidth}
    \centering
    \includegraphics[width=\linewidth]{\detokenize{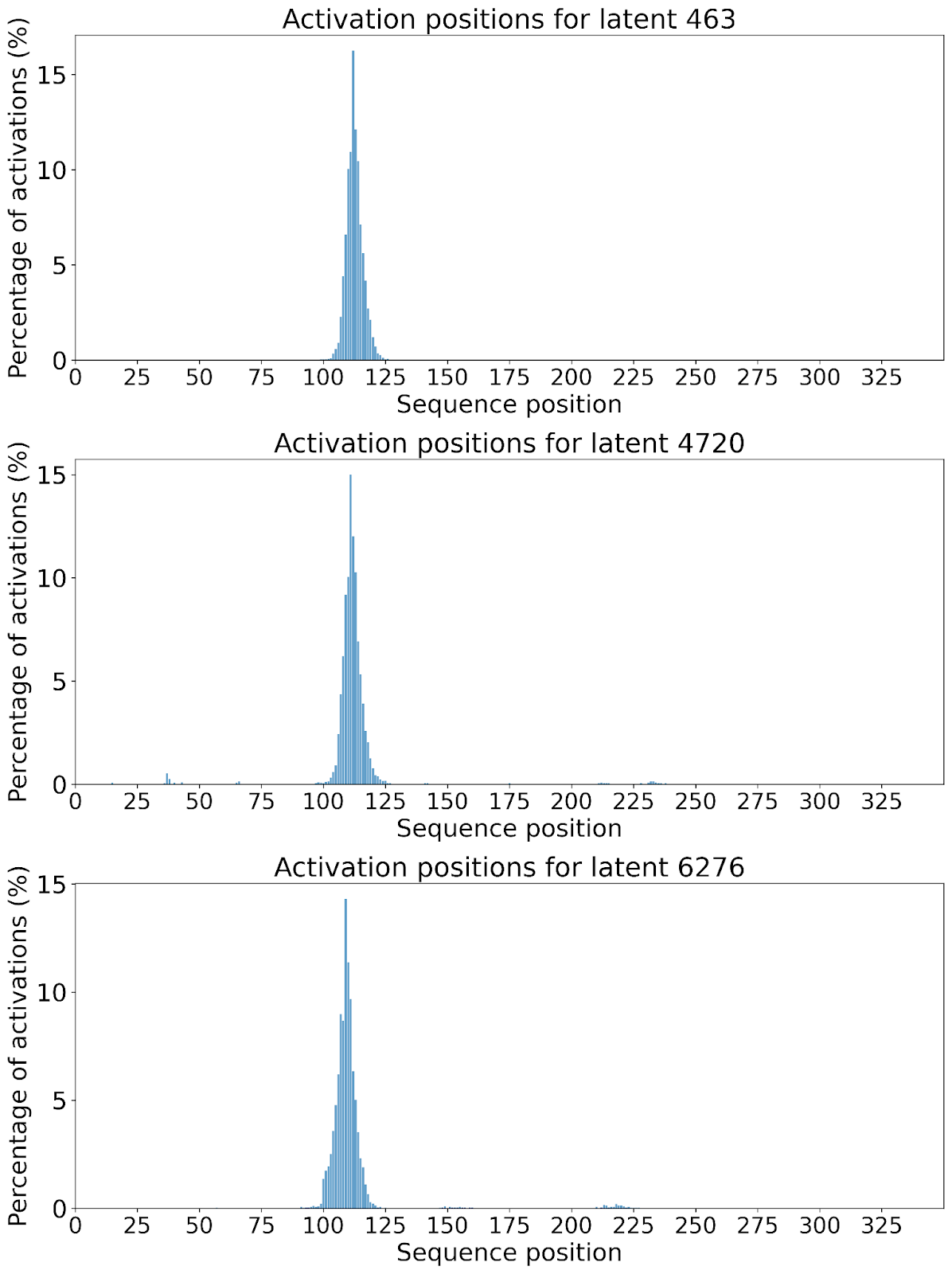}}
    \subcaption{Absolute positional activations}
    \label{fig:IMGT_absolute}
  \end{subfigure}
  \hfill
  \begin{subfigure}[t]{0.48\textwidth}
    \centering
    \includegraphics[width=\linewidth]{\detokenize{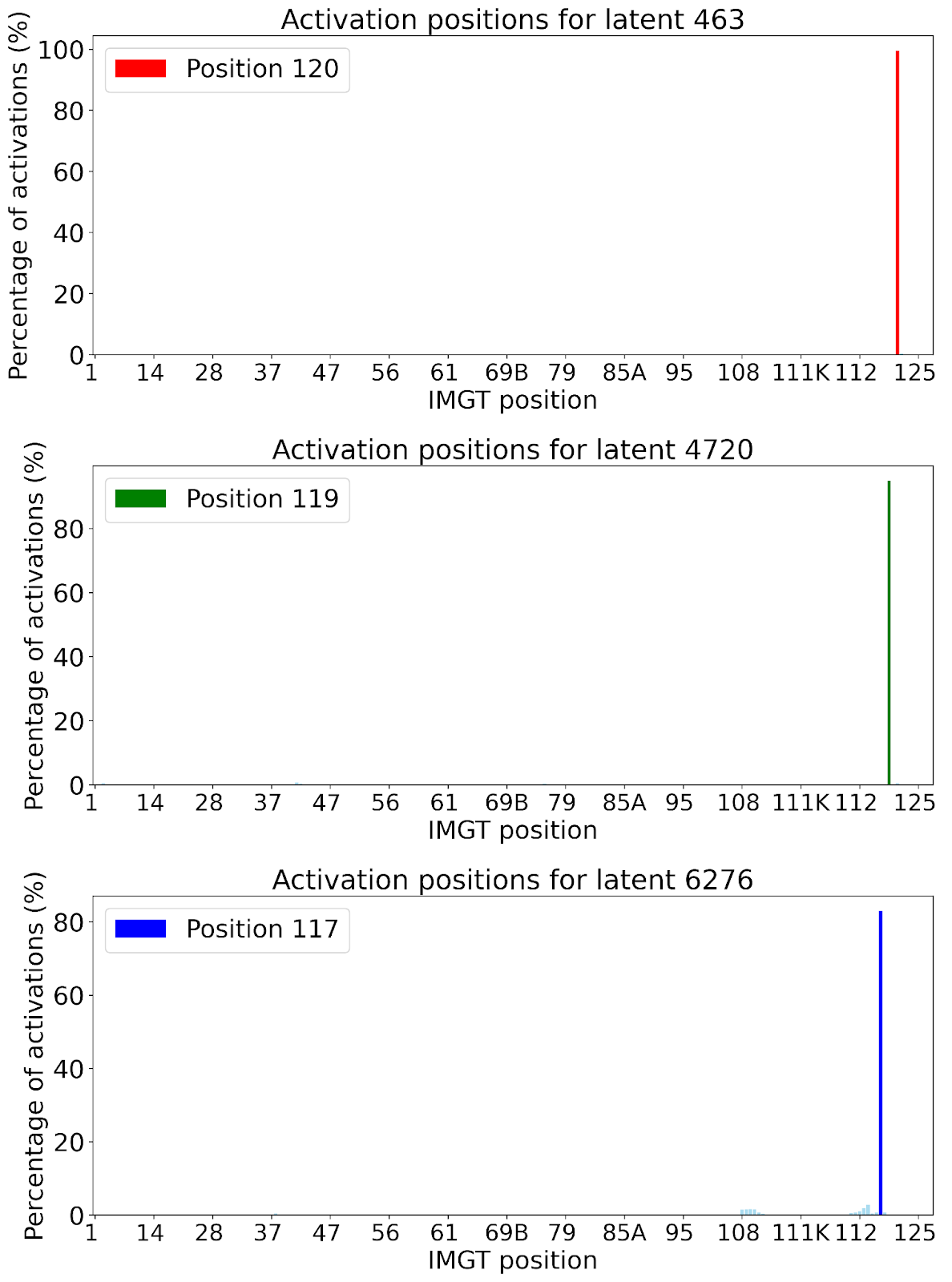}}
    \subcaption{IMGT activations}
    \label{fig:IMGT_imgt}
  \end{subfigure}

  \caption{Comparison of absolute positional (a) and IMGT (b) activations of top three IGHJ4 TopK p-IgGen latents ranked by $F_1$ score. Only one of the three pass the 'feature-threshold' but the other latents still show an $F_1$ score greater than $F_{1,\mathrm{all+}}$. The sequence/IMGT positions are shown on the x-axis. For the sequence positions, the amino acid sequences were end-padded to a constant length of 350. Percentage of total activations on any given position across validation IGHJ4 sequences is shown on the y-axis. The most frequent IMGT position for activation is highlighted for each latent. Latent activations show a distribution near the end of the heavy chain when aligned based on absolute sequence position. In contrast, latents demonstrate discrete activations when aligned based on IMGT numbering.}
  \label{fig:IMGT_IGHJ4_activations}
\end{figure*}

First, we looked at the absolute positional activations of top IGHJ4 TopK latents, based on $F_1$ score, across all the sequences in our validation set which had an IGHJ4 heavy J gene, and compared it to activations on specific IMGT positions. Whilst activations on absolute positions were distributed near the end of the heavy chain corresponding to the J region, the activations on IMGT positions were more consistent and concentrated. This implies that the model does not base its activation pattern on the absolute sequence length alone, but rather the underlying sequence alignment (Figure \ref{fig:IMGT_IGHJ4_activations}). 

\captionsetup{font=small,skip=6pt}

We then chose to investigate the specific residue identities on which the latents were activated. Based on the heavy J gene sequence alignments, the top two latents activated at IMGT positions 120 and 119, which are a Q and G, respectively. These are conserved across all human IGHJ genes. The third top latent activated on Y at position 117, which is unique for IGHJ4 \cite{scaviner_protein_1999}. These results indicated that top latents encoded contextual information of the preceding residues. 

\begin{figure*}[t]
  \includegraphics[width=\textwidth]{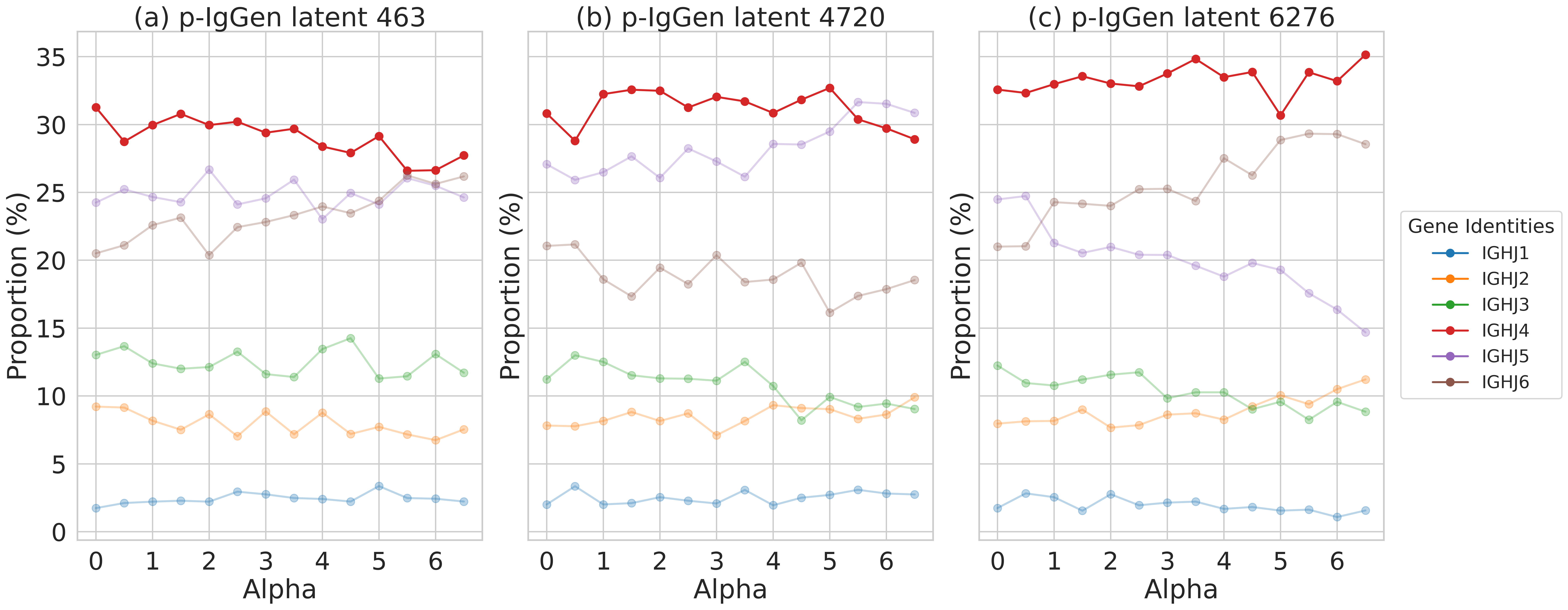}
    \caption{Results of IGHJ4 feature steering for TopK p-IgGen latent 463 (a), 4720 (b), 6276 (c). Y-axis shows the proportion of generated sequences. Plots are coloured by heavy J gene identity. X-axis shows the steering factor used (alpha). Results are for a library of 1000 p-IgGen-generated sequences. For each latent tested (a-c), steering did not result in a predictable change in library composition.}
    \label{fig:Unsuccessful steering}
\end{figure*}

Previous studies have highlighted how highly correlated features may be used to steer model outputs \cite{templeton_scaling_2024, simon_interplm_2024}. We attempted to steer on each identified latent to investigate how it affects model generation. We positively steered on each latent, which we hypothesised should increase the proportion of IGHJ4 in generated sequences. However,  steering on these latents was unpredictable and did not consistently increase IGHJ4 proportions (Supplementary Figure \ref{fig:Unsuccessful steering}). 

To check if this phenomenon was somehow exclusive for IGHJ4 and layer 3, we attempted to steer across all the layers for a number of different features for various gene identities, but were unable to predictably steer model generation [data not shown]. The lack of steerability may indicate how these features individually do not contribute to the gene identity, making them informative features when used for downstream predictions, but not for biasing model output. 

This may be due to feature splitting \cite{chanin_is_2024} which has been reported for TopK SAEs. Feature splitting refers to the phenomenon where higher-order features are broken down into specific contextual examples. In the case of text-based language models, 'math' may be split into 'algebra' and 'geometry'. These phenomena arise when enforcing sparsity in a dictionary consisting of hierarchical features. In this instance, if the identified latents correspond to only single residues within the J-domain, it essentially becomes a residue-level feature as opposed to a sequence-level feature. If the feature activates on a residue specific to the gene identity, it may be a good predictor for the gene identity, but not a steerable feature. This points to the possibility that several features together confer J gene identity, and that these features are likely correlated to each other. Hence, activating one but not the others does not necessarily result in a predictable shift in model performance.

The case study on IGHJ4 indicated that identified features retain biologically relevant context information. Most highly correlated features (based on LR correlation weight and $F_1$ score) tend to be residue-specific. Targeted approaches such as this cannot easily find abstract, higher-order features, assuming they are represented within the latent space to begin with. Concept-specific targeted feature identification might identify highly correlated features that are biologically informative. For instance, two of the three top features (463 and 4720) activate on conserved residues preceded by sequence motifs specific to the gene identity. The third, Latent 6276 activated on an IGHJ4-specific residue, which may explain why this feature can be used to accurately identify IGHJ4. 

Highly predictive features may be correlated with other biologically informative features. Steering on these features did not produce predictable results, making it difficult to interpret the contribution of each individual latent to model generation. Overall, TopK SAEs can identify features in targeted concept analysis which are intuitively interpretable, however, not necessarily steerable. One of the reasons behind the lack of steerability of TopK latents may stem from feature splitting \cite{chanin_is_2024}, as highlighted earlier. As shown in this paper, a solution may be to use nested architectures which limit feature splitting, allowing greater steerability \cite{chanin_is_2024}.
\newpage
\subsection{Supplementary Figures and Tables}
 
\begin{table*}[ht]
  \centering
  \caption{Layer 3 TopK p-IgGen feature counts and maximum $F_1$-scores for IGHJ genes. A latent is counted as a feature if it satisfies $F_1 > F_{1,\mathrm{all+}} + 0.2$ for any tested activation threshold, where $F_{1,\mathrm{all+}}=\frac{2\pi}{1+\pi}$ and $\pi$ is the validation prevalence of the positive class.}
  \label{tab:ighj-features}
  \begin{small}
  \begin{tabular}{lccc}
    \toprule
    Gene & Number of features & Max $F_1$ & $F_{1,\mathrm{all+}}+0.2$ \\
    \midrule
    IGHJ1 & 1  & 0.366 & 0.244 \\
    IGHJ2 & 6  & 0.758 & 0.393 \\
    IGHJ3 & 17 & 0.866 & 0.364 \\
    IGHJ4 & 1  & 0.752 & 0.713 \\
    IGHJ5 & 0  & 0.486 & 0.598 \\
    IGHJ6 & 16 & 0.866 & 0.517 \\
    \bottomrule
  \end{tabular}
  \end{small}
\end{table*}

\captionsetup{font=small,skip=6pt}

\begin{figure*}[t]
  \centering

  \begin{subfigure}[t]{0.48\textwidth}
    \centering
    \includegraphics[width=\linewidth]{\detokenize{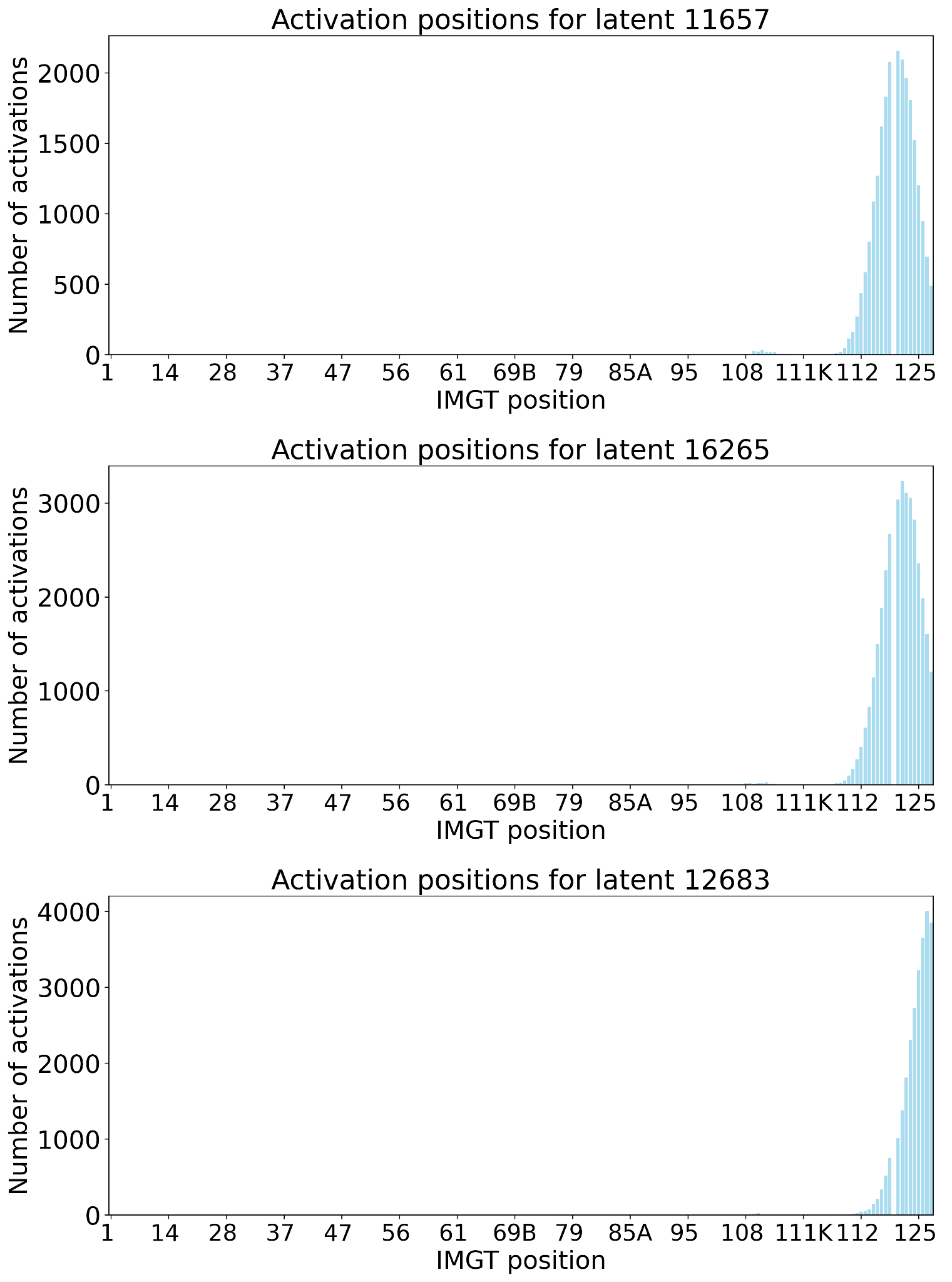}}
    \subcaption{IgLM activations}
    \label{fig:IgLM_activations}
  \end{subfigure}
  \hfill
  \begin{subfigure}[t]{0.48\textwidth}
    \centering
    \includegraphics[width=\linewidth]{\detokenize{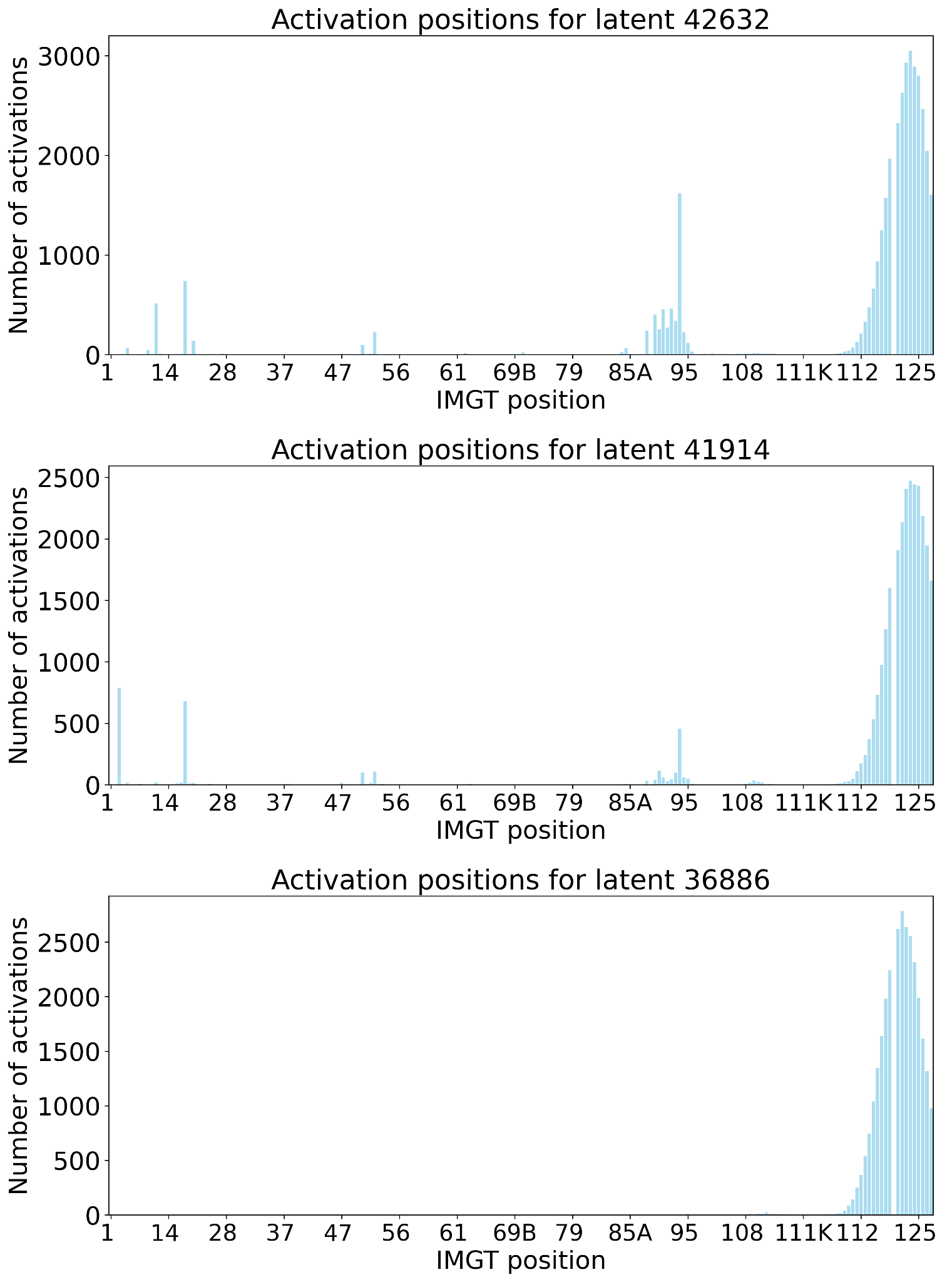}}
    \subcaption{ProGen2-OAS activations}
    \label{fig:ProGen2_activations}
  \end{subfigure}

  \caption{IMGT activations of top three TopK IgLM (a) and ProGen2-OAS (b) IGHJ4 features. The IMGT positions are shown on the x-axis. Percentage of total activations on any given position across validation IGHJ4 sequences is shown on the y-axis. Latent activations show a distribution near the end of the heavy chain when aligned based on IMGT numbering.}
  \label{fig:IgLM_ad_progen2_activations}
\end{figure*}
\captionsetup{font=small,skip=6pt}

\begin{figure*}[!b]
    \centering

    \begin{subfigure}[t]{\textwidth}
        \centering
        \includegraphics[width=\linewidth]{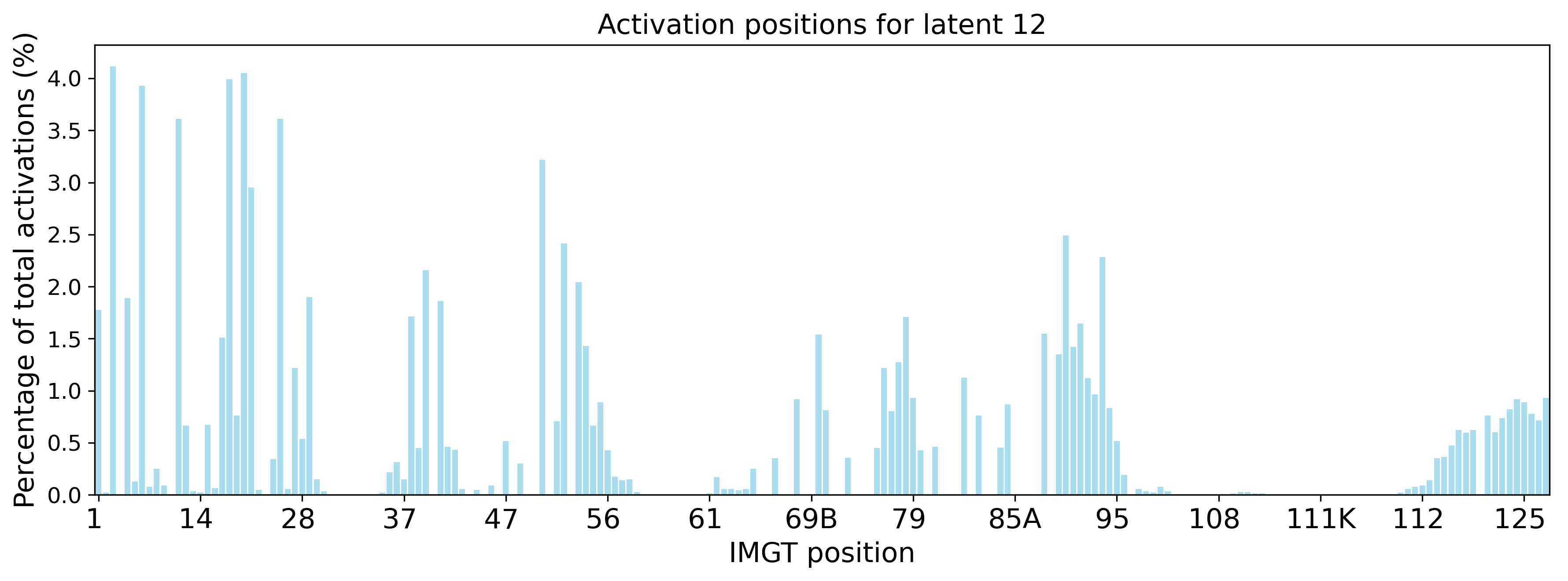}
        \subcaption{Activation pattern of positively correlated O-SAE latent for p-IgGen}
        \label{fig:nested_latent_pos}
    \end{subfigure}

    \vspace{6pt}

    \begin{subfigure}[t]{\textwidth}
        \centering
        \includegraphics[width=\linewidth]{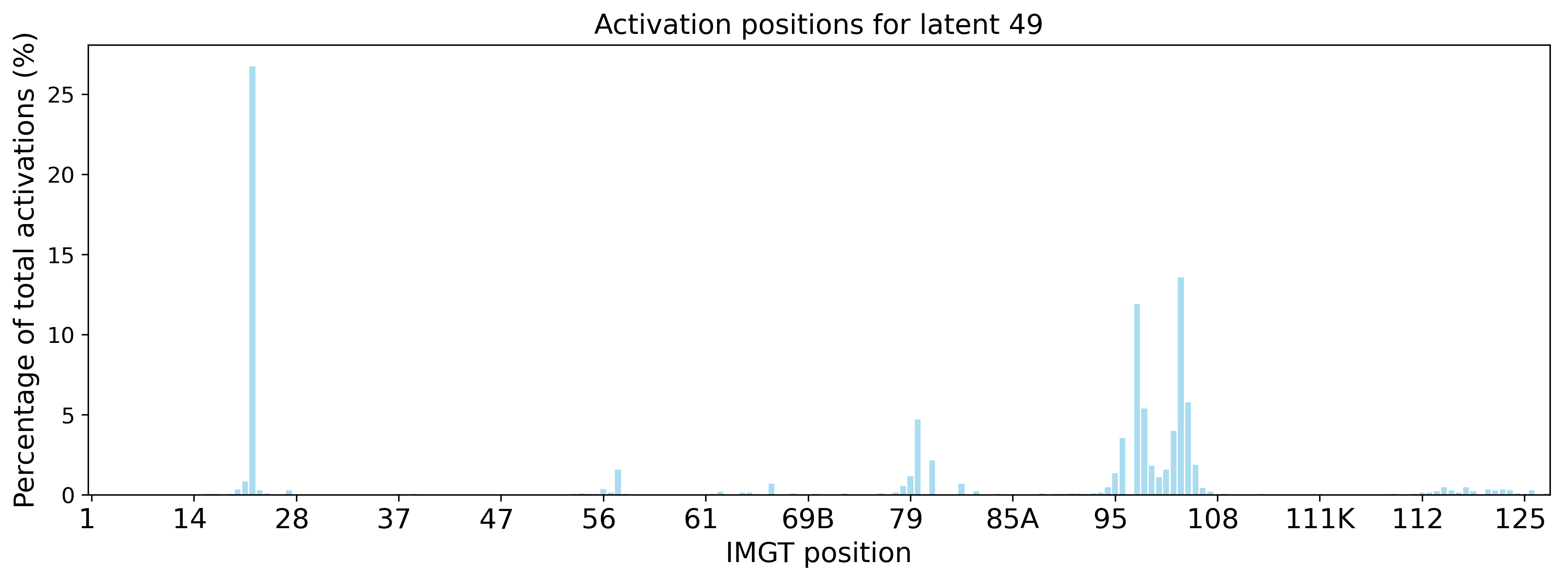}
        \subcaption{Activation pattern of top negatively correlated O-SAE latent for p-IgGen}
        \label{fig:nested_latent_neg}
    \end{subfigure}

    \caption{IMGT activations of p-IgGen O-SAE latent 12 (positively correlated) (a) and 49 (negatively correlated) (b). Activation patterns of both latents show scattered distribution across the range of IMGT positions.}
    \label{fig:nested latent activation}
\end{figure*}


\end{document}